\documentclass[10pt, conference, compsocconf]{IEEEtran}

\usepackage{caption}
\usepackage{subcaption}
\usepackage{multicol, blindtext}

\usepackage{multirow}
\usepackage{textcomp}

\ifCLASSINFOpdf
   \usepackage[pdftex]{graphicx}
   \graphicspath{{}{../jpeg/}}
   \DeclareGraphicsExtensions{.pdf,.jpeg,.png}
\else

\fi

\begin{document}

\title{Improving state-of-the-art in Detecting Student Engagement\\with Resnet and TCN Hybrid Network}

\author{\IEEEauthorblockN{Ali Abedi and Shehroz S. Khan}
\IEEEauthorblockA{KITE, Toronto Rehabilitation Institute\\
University Health Network\\
Toronto, Canada\\
ali.abedi@uhn.ca, shehroz.khan@uhn.ca}
}

\maketitle

\begin{abstract}
Automatic detection of students' engagement in online learning settings is a key element to improve the quality of learning and to deliver personalized learning materials to them. Varying levels of engagement exhibited by students in an online classroom is an affective behavior that takes place over space and time. Therefore, we formulate detecting levels of students' engagement from videos as a spatio-temporal classification problem. In this paper, we present a novel end-to-end Residual Network (ResNet) and Temporal Convolutional Network (TCN) hybrid neural network architecture for students' engagement level detection in videos. The 2D ResNet extracts spatial features from consecutive video frames, and the TCN analyzes the temporal changes in video frames to detect the level of engagement. The spatial and temporal arms of the hybrid network are jointly trained on raw video frames of a large publicly available students' engagement detection dataset, DAiSEE. We compared our method with several competing students' engagement detection methods on this dataset. The ResNet+TCN architecture outperforms all other studied methods, improves the state-of-the-art engagement level detection accuracy, and sets a new baseline for future research.
\end{abstract}


%
\IEEEpeerreviewmaketitle

\section{Introduction}
With the widespread availability and adoption of internet services across major urban centers and universities, online education, telehealth, telemedicine, and telerehabilitation are becoming more ubiquitous and mainstream. In situations such as the COVID-19 pandemic with strict social distancing guidelines, these online services made it possible for students to complete their courses \cite{mukhtar2020advantages} and patients to receive the necessary care and health support \cite{nuara2021telerehabilitation}. Online services offer many advantages compared to the traditional in-person settings, in terms of being more accessible, economical, and customizable \cite{mukhtar2020advantages}. However, these online services also bring other types of challenges. For instance, in online classroom setting, students and tutor are behind a ‘virtual wall’, it becomes very difficult for the tutor to assess the students engagement in the class being taught \cite{venton2021strategies}. This problem is further exacerbated if the group of students is large \cite{sumer2021multimodal}. Therefore, from a tutor’s perspective, it is important to automatically detect the level of engagement among students to provide them real-time feedback and take necessary actions to engage the students to maximize their learning objectives.

Various modalities have been utilized to automatically detect students' engagement, including students' images \cite{whitehill2014faces}, videos \cite{dhall2020emotiw, gupta2016daisee}, audio \cite{dhall2020emotiw, guhan2020abc}, and Electrocardiogram (ECG) \cite{doherty2018engagement}. Video cameras/webcams are mostly used in student learning environment; thus, they have been extensively used in assessing students' engagement in online classroom setting. Video cameras and webcams offer a cheaper, ubiquitous and unobtrusive alternative to other sensing modalities. Therefore, majority of the recent works on student engagement detection are based on the visual data of students acquired by cameras and using computer-vision techniques \cite{doherty2018engagement, dewan2019engagement}.

The computer-vision based approaches for student engagement are categorized into image-based and video-based approaches. The former approaches detect engagement based on single images \cite{nezami2019automatic}, or single frames extracted from the videos \cite{whitehill2014faces}. A major limitation of this approach is that it only utilizes spatial information from single frames, whereas engagement detection is a spatio-temporal affective behaviour \cite{d2017advanced}. Another challenge with frame based approaches is that annotation is needed for each frame, which is an arduous task in practice. The latter approach is to detect students' engagement from videos instead of using single frames. In this case, one label is needed after each video segment. Less annotations is required in this case; however, the classification problem is more challenging due to the coarse labeling.

In traditional settings, in the video-based approaches, handcrafted features, such as eye gaze and head pose, can be extracted and classification algorithms can be trained to detect the level of engagement \cite{gupta2016daisee,huang2019fine,wang2020automated,geng2019learning,zhang2019novel}. More recently, end-to-end video-based approaches have been proposed to detect student engagement, in which consecutive raw frames of video are fed to variants of Convolutional Neural Networks (CNNs) to detect the level of engagement \cite{gupta2016daisee, huang2019fine, wang2020automated, geng2019learning, zhang2019novel}.

Sinatra et al. \cite{sinatra2015challenges} defined students' engagement from person-oriented perspective: engagement is the cognitive, affective, and motivational \emph{states} of the student at the moment of learning and are best captured with fine-grained physiological and behavioral measures (e.g., facial expressions, actions, body postures). D'Mello et al. \cite{d2017advanced} implied that students' engagement is not stable over time, and should be assessed at fine-grained time scales ranging from seconds to a few minutes. For this reason, we consider students' engagement as a spatio-temporal data (video) analysis problem. To detect the level of engagement from video data, in addition to the \emph{state} of the student in each video frame, the changes in the \emph{state} of the student over consecutive frames need to be analyzed. Therefore, for detecting students' engagement in an end-to-end setting, utilizing sequential neural networks in conjunction with CNNs can be beneficial. With this motivation, we propose a novel end-to-end Residual Network (ResNet) \cite{szegedy2015going} and Temporal Convolutional Network (TCN) \cite{bai2018empirical} hybrid neural network architecture (Resnet+TCN) for detecting students' engagement level from videos. The ResNet extracts spatial features from consecutive frames, and the TCN analyzes the temporal changes in consecutive frames to detect the level of engagement. We evaluate the performance of the Resnet+TCN architecture on a publicly available dataset, the Dataset for the Affective States in E-Environments (DAiSEE) \cite{gupta2016daisee}. The DAiSEE dataset contains 9,068 10-second videos captured from 112 students in online classroom setting, and annotated by the  engagement level of students. Our results on the DAiSEE dataset outperformed other competing methods and improved the results on the state-of-the-art methods by 3.9\% in engagement level detection accuracy.

\section{Literature Review}
\label{review}
Over the recent years, extensive research efforts have been devoted to study student engagement detection using computer vision and deep learning techniques \cite{doherty2018engagement, dewan2019engagement, dhall2020emotiw}. We review recent works on the computer-vision based student’s engagement detection with a focus on the works that performed their experiments on the DAiSEE dataset (described in Section. \ref{results}).

Two types of approaches were found prominently in computer-vision based engagement detection problem, i.e., feature-based models and end-to-end models.

In the \textbf{feature-based} engagement detection approaches, firstly, multi-modal handcrafted features are extracted from video/image and then fed to a classifier or regressor to detect the level of engagement in video/image \cite{dewan2019engagement, dhall2020emotiw, wu2020advanced, zhu2020multi}. Wu et al. in \cite{wu2020advanced} proposed a feature-based approach for student's engagement level detection in EmotiW dataset \cite{dhall2020emotiw}. They extracted facial and upper-body features from videos and classified the features using a combination of Long Short-Term Memory (LSTM) and Gated Recurrent Unit (GRU) to detect the level of engagement. Zhu et al. in \cite{zhu2020multi} proposed an attention-based GRU model to classify hand-crafted face and body features from videos and detect the level of engagement in the EmotiW dataset \cite{dhall2020emotiw}. Whitehill et al. in \cite{whitehill2014faces} proposed different feature-extraction (box filters and  Gabor features) and classification (SVM and GentleBoost) combinations to detect the level of engagement of students from single images in their dataset.

Huang et al. \cite{huang2019fine} proposed Deep Engagement Recognition Network (DERN) which combines bidirectional LSTM and attention mechanism to classify extracted features from faces and detect the level of engagement. They achieved 60\% engagement level detection accuracy on the DAiSEE dataset. Wang et al. \cite{wang2020automated} proposed a CNN architecture to classify facial landmarks and features extracted from faces to detect the level of engagement with an accuracy of 57\% on the DAiSEE dataset.

In the \textbf{end-to-end} approaches, the raw frames of videos or images are fed to a deep CNN classifier/regressor to detect engagement. Gupta et al. \cite{gupta2016daisee} introduced the DAiSEE dataset  and established benchmark results using different end-to-end convolutional video-classification techniques, including the InceptionNet \cite{szegedy2015going}, C3D \cite{tran2015learning}, and Long-term Recurrent Convolutional Networks (LRCN) \cite{donahue2015long}, achieving 46.4\%, 56.1\%, and 57.9\% accuracy, respectively.

Geng et al. \cite{geng2019learning} utilized C3D classifier along with the focal loss to classify the level of engagement in the DAiSEE dataset, and achieved 56.2\% accuracy. Zhang et al. \cite{zhang2019novel} proposed a modified version of the Inflated 3D (I3D) along with the weighted cross entropy loss to classify the level of engagement in the DAiSEE dataset, and achieved 52.35\% accuracy. In addition to the original four-level engagement classification problem in the DAiSEE, Zhang et al. \cite{zhang2019novel} considered the engagement detection in the DAiSEE as a two-class classification problem. They changed the labels from low and very-low levels of engagement to not-engaged and from high and very-high levels of engagement to engaged. They trained the I3D model to solve the binary, not-engaged/engaged, classification problem, and obtained 98.82\% accuracy. The above two loss functions, the focal loss in \cite{geng2019learning} and the weighted cross entropy loss in \cite{zhang2019novel}, are used to tackle the problem of imbalanced data distribution in the DAiSEE dataset.

Liao et el. \cite{liao2021deep} proposed Deep Facial Spatio-Temporal Network (DFSTN) for students' engagement detection in online learning. Their model for engagement detection contains two modules, a pretrained SE-ResNet-50 is used for extracting spatial features from faces, and a LSTM with global attention for generating an attentional hidden state. They evaluated their method on the DAiSEE dataset and achieved an accuracy of 58.84\%.

Dewan et el. \cite{dewan2018deep} and \cite{murshed2019engagement}, modified the original four-level video engagement annotations in the DAiSEE dataset, and defined two and three-level engagement detection problems based on the labels of other emotional states in the DAiSEE dataset. They also changed the video engagement detection problem in the DAiSEE dataset to image engagement detection problem and performed their experiments on 1800 images extracted from videos in the DAiSEE dataset. For the two-level engagement detection (not-engaged and engaged), the face images originally labeled as bored, confused, and frustrated are assigned the label not-engaged. The labels for the engaged face images has not been changed. For the three-level engagement detection (not-engaged, normally-engaged, and very-engaged), the engaged face images with intensity 1 and 2 are given the label normally-engaged. The engaged face images with the intensity values higher than 2 are given the label very-engaged. The face images with labels bored, confused, and frustrated are given the label not-engaged \cite{dewan2018deep}. In \cite{dewan2018deep}, they used Local Directional Pattern (LDP) to extract person-independent edge features for different facial expressions, Kernel Principal Component Analysis (KPCA) to capture the nonlinear correlations among the extracted features, and Deep Belief Networks (DBN) to classify the extracted features and detect the level of engagement. They achieved 90.89\%, and 87.25\% accuracy for two and three-level engagement detection, respectively. In \cite{murshed2019engagement}, the extracted face regions are fed to different 2D CNN architectures to detect the level of engagement, and achieved 92.33\% accuracy for two-level engagement detection. The authors in \cite{dewan2018deep} and \cite{murshed2019engagement} have altered the original \emph{video} engagement detection problem in the DAiSEE dataset \cite{gupta2016daisee} to \emph{image} engagement detection problem and have not evaluated their methods on the video test set in the DAiSEE dataset (described in Section \ref{results}). Therefore, their reported accuracy would be hard to verify for generalization of results and their methods working with \emph{images} cannot be compared to other works on the DAiSEE dataset that use \emph{videos}.

We formulate the student engagement problem from videos as a spatio-temporal classification problem. We first extract spatial features from raw frames using a ResNet architecture. Since the engagement behavior varies across different video frames, we build a TCN network on the spatial features to model the temporal variation. Next, we describe the ResNet+TCN architecture.

\begin{figure*}[ht]
    \centering
    \includegraphics[scale=0.36]{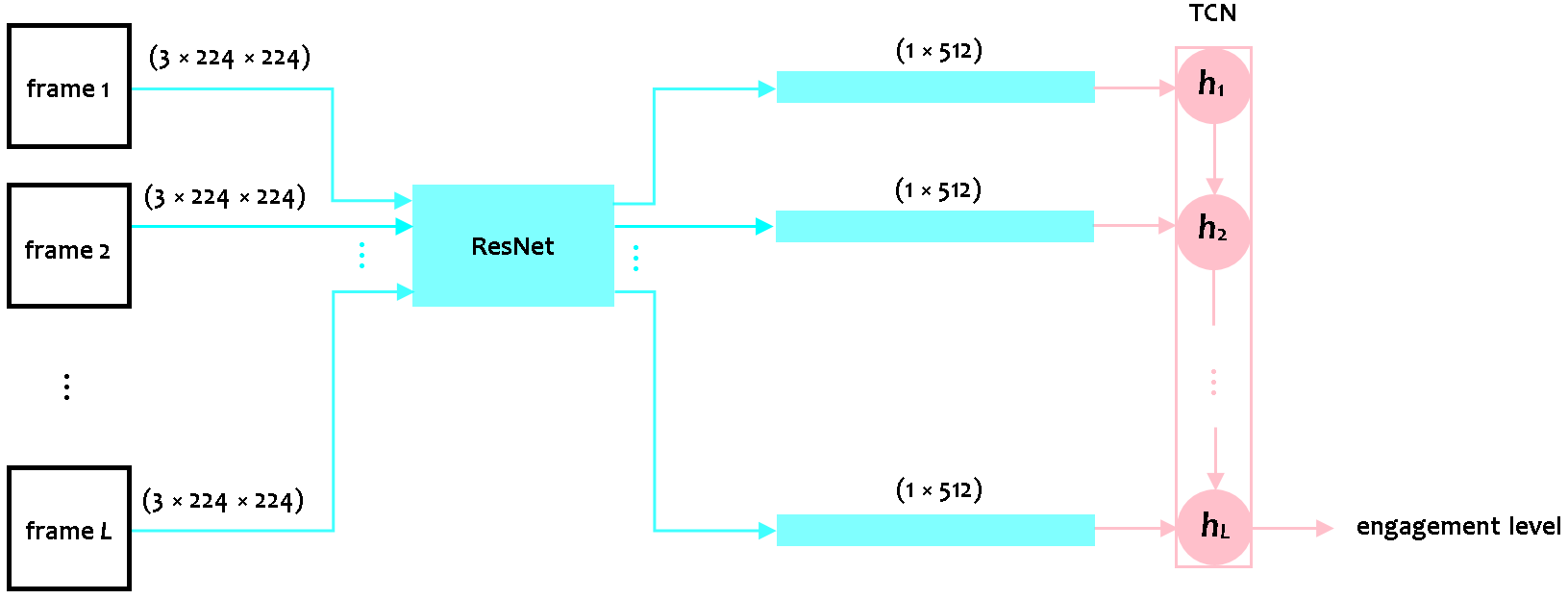} \\
    \caption{The proposed end-to-end architecture, ResNet+TCN, for engagement level detection from video. The input to the ResNet+TCN is a sequence of \emph{L} raw frames of video, and the output is detected ordinal class corresponding to the level of engagement of student in the video. The 2D ResNet extracts spatial features from consecutive video frames, and the TCN analyzes the temporal changes in video frames to detect the level of engagement. The extracted feature vectors (by ResNet) from the consecutive frames are considered as the input to the consecutive time steps of the TCN. One fully connected layer after the last time step of the TCN outputs the predicted class.}
    \label{fig:architecture}
\end{figure*}

\section{ResNet+TCN Architecture for Detecting Student Engagement}
\label{proposed}
Figure. \ref{fig:architecture} shows the structure of the ResNet+TCN hybrid neural network architecture for determining student engagement levels in videos. This method is end-to-end, without the need to extract manual or handcrafted features from videos or frames, and the features are learned on the fly while jointly training the network. The input to the ResNet+TCN is a sequence of raw frames of video, and the output is detected ordinal class corresponding to the level of engagement of student in the video.

As discussed above, engagement is a spatio-temporal affective state that takes place in consecutive frames of video over time, its detection requires analyzing video spatially and temporally. The ResNet and TCN are powerful deep neural network architectures for spatial and temporal analysis of large volumes of data. In previous works, spatial and temporal neural network architectures have been combined in different video-analysis applications. For instance, Donahue et al. \cite{donahue2015long} combined 2D CNN and LSTM and proposed LRCN for activity recognition and description in video. Petridis et al. \cite{petridis2018end} combined 3D CNN and ResNet (for spatial feature extraction from video) with bidirectional GRU (for temporal analysis) for lip reading in video. Ma et al. \cite{ma2021lip} modified the method proposed by Petridis et al. \cite{petridis2018end} by substituting the GRU with the TCN \cite{bai2018empirical} for lip reading in video.

We adapt an altered version of the architecture proposed in \cite{ma2021lip} and propose a new architecture using ResNet and TCN. In \cite{ma2021lip}, spatio-temporal features are extracted from videos using 3D CNN. The extracted features by 3D CNN is given to ResNet, and the output of ResNet is given to TCN for temporal analysis. In our architecture, spatial features are extracted from frames using ResNet, and the outputs of ResNet are given to TCN for temporal analysis.

In our proposed architecture, the ResNet \cite{szegedy2015going} and TCN \cite{bai2018empirical} are combined to model and jointly train on the spatio-temporal data in sequences of frames in video. We choose TCN because of its superiority in modeling sequences of larger length and retaining memory of history in comparison to generic recurrent architectures such as LSTMs and GRUs \cite{bai2018empirical}. The ResNet extracts spatial features from single frames, and the TCN models the temporal changes in sequence of frames and outputs the detected level of engagement. Given a video sequence, the input to the ResNet+TCN is an \emph{L} × \emph{C} × \emph{H} × \emph{W} tensor, where \emph{L}, \emph{C}, \emph{H},  and \emph{W} correspond to the number of frames, number of channels, frame height, and frame width, respectively. After removing the final fully-connected layer of the standard Resnet18 \cite{paszke2019pytorch}, it is used as the (trainable) feature extractor from single frames of the input video. The extracted feature vectors from the consecutive frames are considered as the multi-dimensional input to the consecutive time steps of a dilated TCN \cite{bai2018empirical} to model temporal information in the video. The output of the final time step of  the TCN is fed to a fully-connected layer and a softmax function to detect the level of student engagement from the input video.

One common problem in the existing student engagement detection datasets is the highly imbalanced distribution of engagement levels \cite{gupta2016daisee, dhall2020emotiw, whitehill2014faces}. The number of samples with low levels of engagement is much less than the number of samples with high levels of engagement. In such sample distribution, it is highly likely that most of the minority-level samples are classified to the majority levels of engagement. To tackle this problem, we use weighted cross entropy loss function \cite{paszke2019pytorch}. The weights of the cross entropy loss function for each class are determined according to the proportion of the samples in the corresponding class to all the samples in the training set.

Another problem encountered during training the network is that due to random sampling in each training iteration, the batches may not contain samples from all the classes because the minority class samples may be left out. Since, in most of training iterations, there are no samples of the minority classes in training batches, the weighted loss function will not be able to affect the network to be trained by the minority class samples. To circumvent this problem, we adopted a customized sampling strategy in which the samples of all classes are included in each batch of the Stochastic Gradient Descent (SGD) during training \cite{paszke2019pytorch}.

\begin{table}[h]
\begin{center}
 \begin{tabular}{||c c c c||} 
 \hline
 level & train & validation & test \\ [0.5ex]
 \hline\hline
 0 & 34 & 23 & 4 \\ 
 \hline
 1 & 213 & 143 & 84 \\
 \hline
 2 & 2617 & 813 & 882 \\
 \hline
 3 & 2494 & 450 & 814 \\
 \hline
 total & 5358 & 1429 & 1784 \\
 \hline
\end{tabular}
\end{center}
\vspace{0.2cm}
\caption{The number of samples in train, validation, and test sets in different levels of engagements in the DAiSEE dataset \cite{gupta2016daisee}, see Section. \ref{dataset}.}
\label{tab:distribution}
\end{table}

\section{Experimental Results}
\label{results}

\subsection{Dataset}
\label{dataset}
The DAiSEE dataset, introduced by Gupta et al. \cite{gupta2016daisee}, is used to evaluate the performance of the ResNet+TCN in comparison to the previous related methods. The dataset contains 9,068 videos captured from 112 students in online courses for recognizing their affective states of boredom, confusion, engagement, and frustration in the wild. We only focus on the student engagement, with four levels of engagement as level 0 (very low), 1 (low), 2 (high), and 3 (very high). These scores were given based on Whitehill et al. \cite{whitehill2014faces}. The length, frame rate, and resolution of the videos are 10 seconds, 30 fps, and 640 × 480 pixels. Table. \ref{tab:distribution} shows the distribution of samples in train, validation, and test sets according to the authors of the DAiSEE dataset \cite{gupta2016daisee}. We use these sets in our experiments to fairly compare the ResNet+TCN method with the previous methods on the DAiSEE dataset. We combined 5358 and 1429 videos in the train and validation sets to train ResNet+TCN model and report results on 1784 test videos. As can be seen in Table. \ref{tab:distribution}, the dataset is highly imbalanced, only 0.63\%, 1.61\%, and 0.22\% percent of train, validation, and test sets are in the level 0 (corresponding to very low engagement).

\subsection{Experimental Setting}
\label{setting}
The videos were down-sampled, temporally and spatially, to get  50 × 3 × 224 × 224 (\emph{L} × \emph{C} × \emph{H} × \emph{W}) tensors as inputs to the ResNet+TCN architecture (as discussed in Section. \ref{proposed}).  The 3 × 224 × 224 dimensions are the standard dimension for input to the ResNet18 \cite{paszke2019pytorch}. The SGD is used for parameter optimization with learning rate and batch size of 0.001 and 5, respectively. The ResNet extracts feature vectors of dimension 512 from the consecutive frames and feeds them to the TCN. The parameters of the TCN, giving the best results, are as follows, 8, 128, 7, and 0.25 for the number of levels, number of hidden units, kernel size, and dropout \cite{bai2018empirical}. We implemented the experiments in PyTorch \cite{paszke2019pytorch} on a server with 64 GB of RAM and NVIDIA Tesla P100 PCIe 12 GB GPU. The code of Resnet+TCN is available at https://github.com/abedicodes/ResNet-TCN.

\subsection{Results}
\label{_results}
\begin{figure}
    \centering
    \includegraphics[scale=0.35]{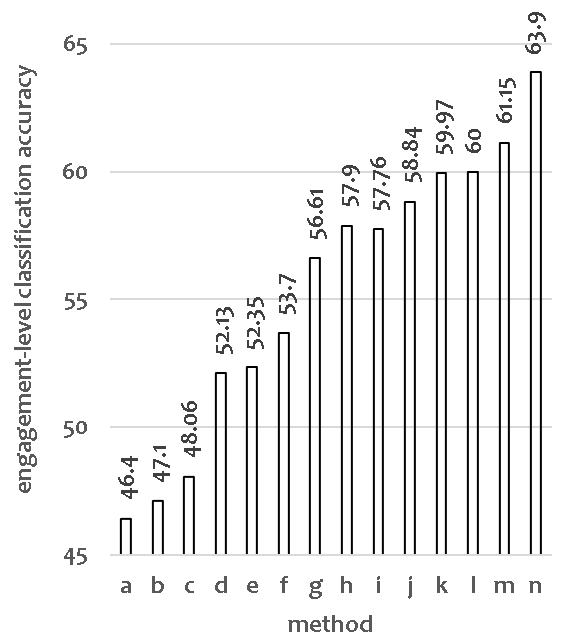} \\
    \caption{Engagement-level classification accuracy of different methods on the DAiSEE dataset \cite{gupta2016daisee},
    (a) video-level InceptionNet \cite{gupta2016daisee},
    (b) frame-level InceptionNet \cite{gupta2016daisee},
    (c) C3D feature extraction \cite{gupta2016daisee},
    (d) C3D averaging + LSTM \cite{parmar2017learning},
    (e) I3D \cite{zhang2019novel},
    (f) ResNet + TCN with sampling and weighted loss (proposed),
    (g) C3D + LSTM \cite{parmar2017learning},
    (h) LRCN \cite{donahue2015long},
    (i) C3D fine tuning \cite{tran2015learning},
    (j) DFSTN \cite{liao2021deep},
    (k) C3D + TCN (proposed),
    (l) DERN \cite{huang2019fine},
    (m) ResNet + LSTM (proposed),
    (n) ResNet + TCN (proposed).}
    \label{fig:accuracy}
\end{figure}

For comparing the ResNet+TCN network with other works, we took reported results from the following methods: video-level and frame-level InceptionNet \cite{gupta2016daisee}, C3D \cite{gupta2016daisee}, I3D \cite{zhang2019novel}, DERN \cite{huang2019fine}, and DFSTN \cite{liao2021deep}. In addition, we implemented the combination of the ResNet with LSTM, and C3D (up to the layer pool-5) \cite{tran2015learning} with LSTM (one-layer unidirectional with 128 hidden neurons) to investigate their performance compared to the ResNet+TCN method. Figure. \ref{fig:accuracy} shows the results of applying different end-to-end methods to the four-class engagement level detection problem in the DAiSEE dataset. In this figure, the accuracy is shown for the previous works and the methods we implemented to evaluate on the DAiSEE dataset. Figure. \ref{fig:accuracy} shows that the ResNet+TCN method achieves the highest accuracy of 63.9\%, which is 2.75\% higher than the best performing method of Resnet with LSTM (61.15\%) and 3.9\% higher than the state-of-the-art DERN method (60\%), and higher than other previous methods. These results show the superiority of the ResNet+TCN in modeling students' engagement level as a spatio-temporal classification problem.

\begin{figure*}
\centering
\includegraphics[scale=0.32]{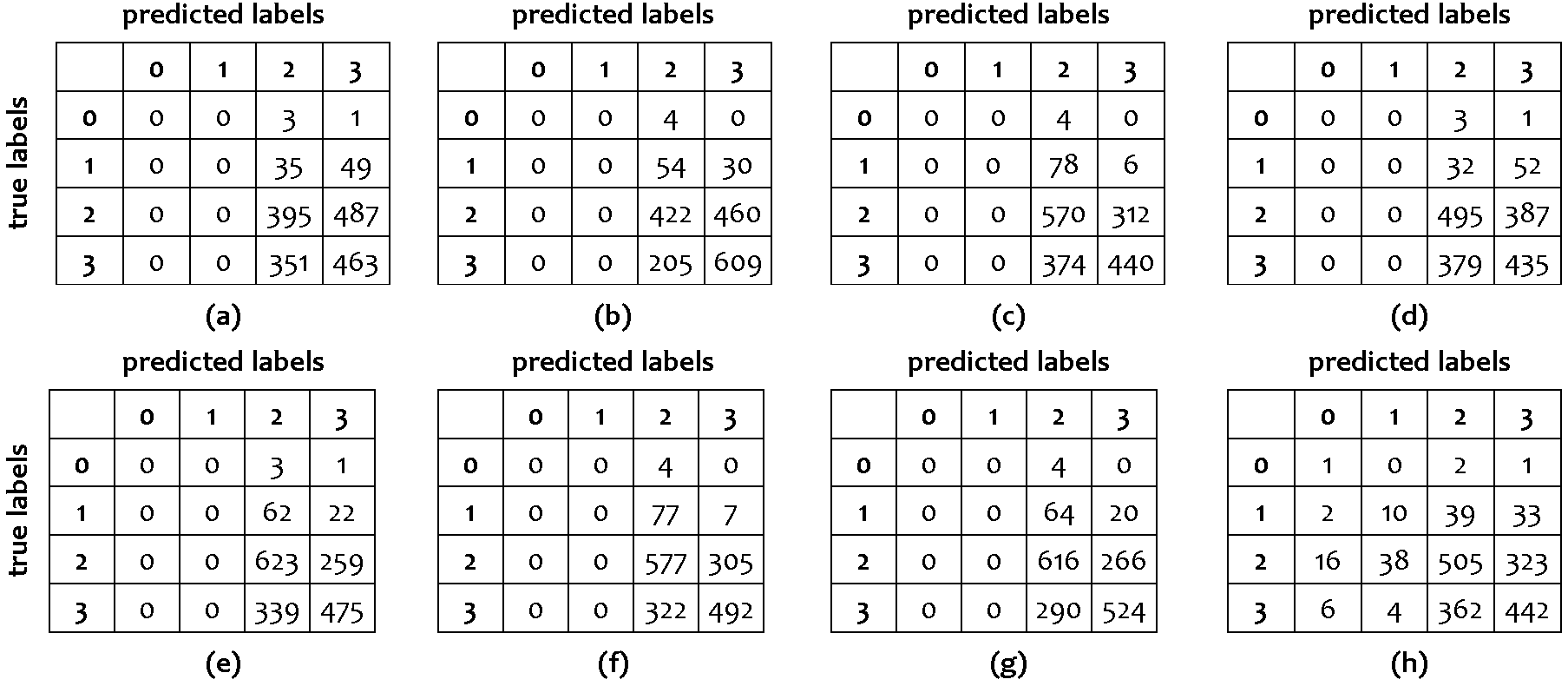} \\     

\caption{Engagement-level confusion matrices of different methods on the DAiSEE dataset \cite{gupta2016daisee},
(a) C3D feature extraction \cite{tran2015learning},
(b) C3D fine tuning \cite{tran2015learning},
(c) C3D + LSTM \cite{parmar2017learning},
(d) C3D averaging + LSTM \cite{parmar2017learning},
(e) ResNet + LSTM (proposed),
(f) C3D + TCN (proposed),
(g) ResNet + TCN (proposed),
(h) ResNet + TCN with weighted sampling and weighted loss (proposed).}
\label{fig:confusion}
\end{figure*}

None of the previous works on the DAiSEE dataset, working with the original four-class annotations, reported their confusion matrices for test set \cite{gupta2016daisee,zhang2019novel,wang2020automated,huang2019fine,geng2019learning,liao2021deep}, and only reported the accuracy results. Therefore, it is hard to determine the individual performance of their methods on each of the engagement levels. We implemented some of the previous methods, including C3D (feature extraction) \cite{gupta2016daisee}, C3D (fine tuning) \cite{gupta2016daisee}, C3D + LSTM \cite{parmar2017learning}, and C3D averaging + LSTM \cite{parmar2017learning} to investigate their confusion matrices compared to the proposed architectures. Figure. \ref{fig:confusion} shows the confusion matrices of the ResNet+TCN compared to the other implemented methods. It can be observed that due to the highly-imbalanced sample distribution, none of the methods are able to classify any samples to the first two classes correctly.

Liao et al. \cite{liao2021deep} reported their confusion matrix for one fold of cross validation during training on the DAiSEE dataset. They didn't report the confusion matrix on the test set of the DAiSEE dataset. Based on their report, they also observed that all the samples were classified to the two higher levels of engagement, and their method cannot correctly classify any samples to the two lower levels of engagement.

As described in Section. \ref{proposed}, after using the customized sampling strategy and weighted loss function in our proposed ResNet+TCN architecture, some samples are correctly classified to the first two levels of engagement (Figure. \ref{fig:confusion} (h)), albeit at the cost of reducing the overall accuracy.

\section{Conclusion and Future Work}
\label{conclusion}
In this paper, we presented a new end-to-end spatio-temporal hybrid architecture, ResNet+TCN, for determining the level of engagement among students in an online classroom setting. We evaluated the performance of the ResNet+TCN method and compared to several previous end-to-end methods. The Resnet+TCN architecture showed improved results in comparison to the state-of-the-art engagement classification accuracy on the DAiSEE dataset. Our results showed that it is very challenging to detect the minority engagement level with very few samples in a supervised classification setting. When class weight term is added to the loss function, some minority level samples were detected at the cost of more false alarms and reducing the overall accuracy of the classifier. In future, we aim to evaluate the performance of the proposed method on other publicly available user engagement datasets, including \cite{dhall2020emotiw}. We aim to work on utilizing end-to-end models on extracted body/face regions of videos. We also aim to investigate the effectiveness of different features \cite{dewan2019engagement,dhall2020emotiw,yang2018deep} in engagement detection and develop neural network models to detect engagement using those features. One possible direction of research to detect the minority levels of engagement is to apply anomaly detection approaches based on autoencoders and generative adversarial networks.

\nocite{}

\bibliographystyle{IEEEtran}

\end{document}